\documentclass[10pt,twocolumn,letterpaper]{article}

\usepackage{cvpr}
\usepackage{times}
\usepackage{epsfig}
\usepackage{graphicx}
\usepackage{amsmath}
\usepackage{amssymb}
\usepackage{algorithm}
\usepackage{algpseudocode}
\usepackage{multirow}
\usepackage{bm}
\usepackage{caption}
\usepackage{subfigure}
\DeclareGraphicsExtensions{.pdf}
\usepackage{indentfirst}

\usepackage[breaklinks=true,bookmarks=false]{hyperref}

\cvprfinalcopy

\def\cvprPaperID{}

\begin{document}

\title{Neural Architecture Search for Deep Face Recognition}

\author{Ning Zhu\\
{\tt\small xyzhuning@126.com}}
\maketitle
\begin{abstract}
By the widespread popularity of electronic devices, the emergence of biometric technology has brought significant convenience to user authentication compared with the traditional password and mode unlocking. Among many biological characteristics, the face is a universal and irreplaceable feature that does not need too much cooperation and can significantly improve the user's experience at the same time. Face recognition is one of the main functions of electronic equipment propaganda. Hence it's virtually worth researching in computer vision. Previous work in this field has focused on two directions: converting loss function to improve recognition accuracy in traditional deep convolution neural networks (Resnet); combining the latest loss function with the lightweight system (MobileNet) to reduce network size at the minimal expense of accuracy. But none of these has changed the network structure. With the development of AutoML, neural architecture search (NAS) has shown excellent performance in the benchmark of image classification. In this paper, we integrate NAS technology into face recognition to customize a more suitable network. We quote the framework of neural architecture search which trains child and controller network alternately. At the same time, we mutate NAS by incorporating evaluation latency into rewards of reinforcement learning and utilize policy gradient algorithm to search the architecture automatically with the most classical cross-entropy loss. The network architectures we searched out have got state-of-the-art accuracy in the large-scale face dataset, which achieves \textbf{98.77\%} top-1 in MS-Celeb-1M and \textbf{99.89\%} in LFW with relatively small network size. To the best of our knowledge, this proposal is the first attempt to use NAS to solve the problem of Deep Face Recognition and achieve the best results in this domain. 
\end{abstract}
\section{Introduction} 
Face recognition is widely used in many fields, such as video surveillance, public security, face payment, and smart home. The critical problem in face recognition is how to acquire facial features accurately. Face recognition can be divided into the following two categories according to features: based on shallow features such as SIFT\cite{sift}, LBP\cite{lbp}, HOG\cite{hog1,hog2} with aggregating them into global face descriptors through some pooling mechanism; based on deep convolution neural network(DCNN). The main advantage of deep learning method is that a large number of datas can be used to train, so as to learn the robust face representation for the changes in training data. This method does not need to design specific robust features for a different type of intra-class differences (illumination, posture, facial expression, age, etc.), but can be learned from training data. In the training process, bottleneck features representing human face are extracted by DCNN, and then other techniques (such as taking advantage of joint Bayesian or exploiting different loss functions) are used to fine-tune the CNN model. CNN architectures for face recognition have been inspired by those magnificent architectures in the \emph{ImageNet Large-scale Visual Recognition Challenge} (ILSVRC)\cite{Imagenet}. Resnet\cite{Resnet} has become the most preferred choice for many target recognition tasks, including face recognition. Many terminals or customers have favored MobileNet's\cite{MobileNet} excellent performance on lightweight devices with limited computing capability. Therefore, selecting loss function for training CNN methods for these two network frameworks has become the most active research field in face recognition recently. The development of loss function in advanced face recognition algorithms mainly includes two ideas:
\begin{itemize}
\item {\bfseries Metric Learning}: Contrastive Loss, Trplet Loss\cite{facenet} and related sampling methods;
\item {\bfseries Margin Based Classification}: Softmax with Center Loss\cite{centerloss}, Sphere Face\cite{sphereface}, Soft-Margin Loss\cite{softmargin}, AM-Softmax (CosFace)\cite{cosineface} and ArcFace\cite{arcface}.
\end{itemize}
In the past few years, remarkable progress has been made in the field of neural architecture search (NAS)\cite{nas}. The computationally intensive NAS algorithms based on reinforcement learning or evolutionary algorithms have proved their ability to produce models that transcend traditional network designed by human beings. Other methods such as DARTS\cite{darts}, SNAS\cite{snas} and Proxyless NAS\cite{proxyless} have also been significantly developed. These methods accelerate the search process by reducing search space and changing search strategies, but the accuracy is relatively lower. For all we know, most NAS papers have trained the image classification benchmark, for instance, CIFAR-10 and ImageNet. On these benchmarks, the NAS algorithm does achieve extraordinary state-of-the-art performance, despite the exorbitant computational cost.

In this paper, we propose a new idea of combining NAS technology with face recognition and customize the neural network structure for this recognition field through a reinforcement learning algorithm. Considering the influence of network structure size, we join latency in the reward of reinforcement learning in the process of neural network search. The goal of this work is to examine whether the neural architecture search method can automatically design better network structures in the field of face recognition.

To prove the effectiveness of this method, we have carried out a lot of experiments. Through the neural network architecture searched by NAS, we achieve 98.77\% accuracy in extensive data set MS-Celeb-1M\cite{msceleb} and 99.89\% accuracy in classical paired data set LFW \cite{lfw}. These all meet the industry's best standards. In addition to achieving excellent efficiency, we also have extraordinary performance in terms of network size: among all the searched networks, the maximum one is only 19.1 M.

The rest of this paper is organized as follows. In Section  \ref{RelatedWorks}, we introduce the evolution of loss function in face recognition and the overview of neural architecture search. We describe the detail of NAS algorithm in Section \ref{Method} which is based on reinforcement learning. In Section \ref{Experiments}, we actualize various experiments to prove that NAS is superior to traditional face recognition algorithms. Finally, we summarize this paper in Section \ref{Conclusions} and provide some guidance for future research.
\section{Related Works}
\label{RelatedWorks}
Loss function plays an essential role in gradient updating of neural networks. A proper loss function can significantly improve the efficiency of the neural network and achieve better results. At present, the loss function in face recognition has been substantially developed. Now we will briefly introduce several essential loss functions.

Neural architecture search can be divided into many types from two dimensions: search space and search strategy. At the same time, the advantages and disadvantages of NAS are evaluated in terms of acceleration strategy, calculation consumption, parameter numbers, inference latency and so on. Next, we will classify the neural architecture search comprehensively.

\subsection{Evolution of Loss function}
For face recognition tasks, we will introduce five most representative loss functions, namely, Cross-Entropy Loss, Angular-Softmax Loss, Additive Margin Softmax Loss, and ArcFace Loss.

\subsubsection{Cross-Entropy Loss}
Cross-entropy loss is one of the most widely used loss functions in classification scenarios\cite{alexnet, goodfellow}. In face recognition tasks, the cross-entropy loss is an effective method to eliminate outliers.\cite{parkhi, deepid2}
Cross-entropy loss is expressed as:
\begin{equation}
\mathcal{L}=-\frac{1}{N}\sum_{i=1}^{N}\log\frac{e^{W^T_{y_i} x_i+b_{y_i}}}{\sum_{j=1}^{n}e^{W^T_j x_i+b_j}},
\label{softmax}
\end{equation}
where $x_i$ is the $i^{th}$ training sample, $N$ is the number of samples, $W_j$ and ${W}_{y_i}$ are the $j^{th}$ and $y_i^{th}$ column of ${W}$, respectively.

\subsubsection{Angular-Softmax Loss}
Based on softmax, the A-SoftMax loss is proposed to enable CNN to learn angular resolution features.
It is described as:
\begin{equation}
\mathcal{L}=\frac{1}{N}\sum_i-\log\big( \frac{e^{\|\bm{x}_i\|\psi(\theta_{y_i,i})}}{e^{\|\bm{x}_i\|\psi(\theta_{y_i,i})}+
	\sum_{j\neq y_i}e^{\|\bm{x}_i\|\cos(\theta_{j,i})}} \big)
\label{angular}
\end{equation}
where $\thickmuskip=2mu \medmuskip=2mu \psi(\theta_{y_i,i})=(-1)^k\cos(m\theta_{y_i,i})-2k$ for $ \theta_{y_i,i}\in[\frac{k\pi}{m},\frac{(k+1)\pi}{m}]$, $\thickmuskip=2mu k\in[0,m-1]$ and $\thickmuskip=2mu m\geq1$ is an integer that controls the size of the angular margin.

\subsubsection{Additive Margin Softmax Loss}
On account of Angular-Softmax Loss, a general function is added to softmax loss for large margin property. The loss function is defined in the following:
\begin{equation}
\begin{aligned}
\mathcal{L} & = -\frac{1}{N}\sum_{i=1}^N{log\frac{e^{s \cdot  \left(cos\theta_{y_i} - m \right)}}{e^{s \cdot \left(cos\theta_{y_i} - m \right)} + \sum_{j=1,j\neq y_i}^{c}{e^{s \cdot  cos\theta_{j}}}}}\\
\end{aligned}
\label{eq:am-softmax}
\end{equation}
where a hyper-parameter $s$ is used to scale up the cosine values.

\subsubsection{ArcFace Loss}
In view of the above loss functions, a new margin $\cos(\theta+m)$ is proposed, which represents the best geometrical interpretation The ArcFace Loss function using angular margin is formulated as:
\begin{equation}
\mathcal{L} =-\frac{1}{m}\sum_{i=1}^{m}\log\frac{e^{s \cdot (\cos(\theta_{y_i}+m))}}{e^{s \cdot\ (\cos(\theta_{y_i}+m))}+\sum_{j=1,j\neq  y_i}^{n}e^{s \cdot \cos\theta_{j}}},
\label{eq:aml}
\end{equation}
\subsection{Neural Architecture Search} 

With the rapid development of AutoML, neural architecture search technology is also changing with each passing day. Different kinds of NAS are springing up like bamboo shoots after a spring rain. Here we will outline NAS in several various aspects.

In principle, NAS search space defines the network architecture to ensure the rationality and quality of the sampling model in the search process. For some specific tasks, introducing prior knowledge can reduce the search space, but to a certain extent, it also restricts the generation of the network beyond previous experience. The backbone structure of the NAS network includes chain architecture space, multi-branch architecture space, and cell/block-based search space. Its operator space contains convolution, pooling, residual connection, and other topological structures.

The search strategy of NAS defines what algorithm can quickly and accurately find the optimal configurations of network architecture. According to different search strategies, NAS can be divided into the following categories.
	
\begin{itemize}
	\item {\bfseries Algorithms based on reinforcement learning }: NASNet, BlockQNN\cite{blockqnn}, ENAS\cite{enas}, MnasNet\cite{mnas}.
	\item {\bfseries Algorithms based on evolution learning }: Hierarchial\cite{hierarchial}, AmoebaNet\cite{amoebaNet}.
	\item {\bfseries Continuous differentiable algorithms }: DARTS, NAO\cite{nao}, DSO\cite{dso}.
	\item {\bfseries Other search algorithms }: SMASH\cite{smash}, PNAS\cite{pnas}, Auto-Keras\cite{autokeras}, Graph hypernet\cite{hypernet}, Efficient Multi-Scale Architectures\cite{multi}, Proxyless NAS 
\end{itemize}

Because NAS requires a lot of training time and computing capability, the commonly used acceleration strategies are parameter sharing, network morphism, and network pruning. In these NAS algorithms, different tasks and scenarios require various performance evaluation indicators. At present, the primary indicators used to evaluate NAS performance are test accuracy, the number of parameters, inference latency, memory utilization, FLOPS, etc.

\section{Method}
\label{Method}
\begin{figure}[htb]
	\begin{center}
		\includegraphics[scale=0.4]{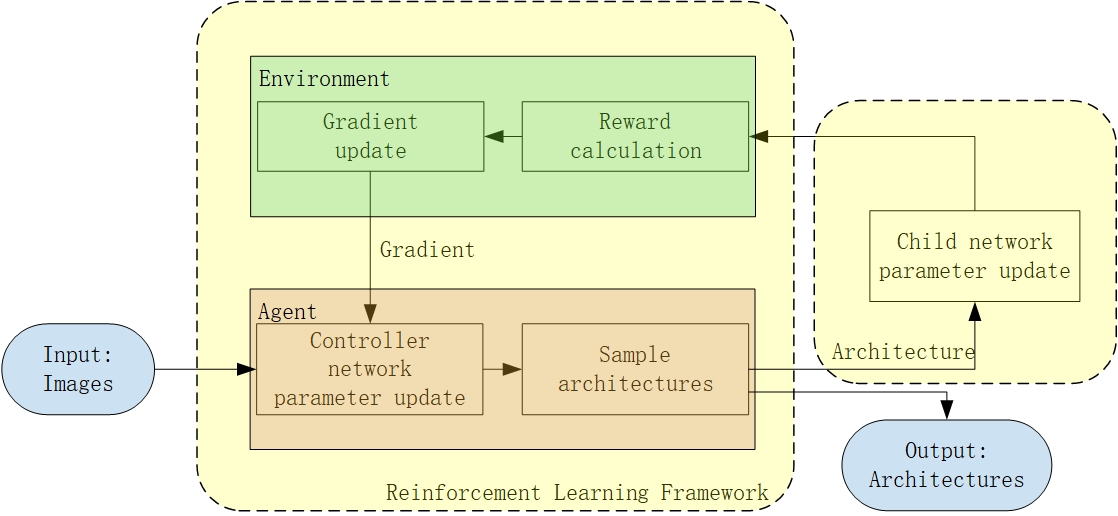}
	\end{center}
	\caption{The overall framework of NAS algorithms. The yellow area on the left is the controller network trained through reinforcement learning; the right field is the training child network}
	\label{nas}
\end{figure}

There are many kinds of NAS, among which the most effective architecture searching methods are those that are computationally intensive. The NAS search space constructed by us consists of all possible directed acyclic graphs (DAGs) on $L$ nodes, in which each node has $N$ arbitrary selected operations. Fig.\ref{nas} visually shows the framework of our algorithm in brief.

In the stage of the searching network, we use reinforcement learning algorithm (policy gradient) to guide the $controller$ network to train the optimal $child$ architecture. The controller network and child network are alternately trained. To improve the efficiency of network search, we refer share child network’s parameters. The controller network is consist of an LSTM which has 100 hidden units. The main trainable parameters are LSTM weights and shared parameters of child network.

The training process of NAS includes two interweaving stages. When training the child network, the generation strategy of the controller network is steadfast. At the same time, when training the controller network, the weight parameters of the child network are fixed — these two stages alternate during NAS training.

\subsection{Training child network}
\begin{figure}[htb]
	\begin{center}
		\includegraphics[scale=0.35]{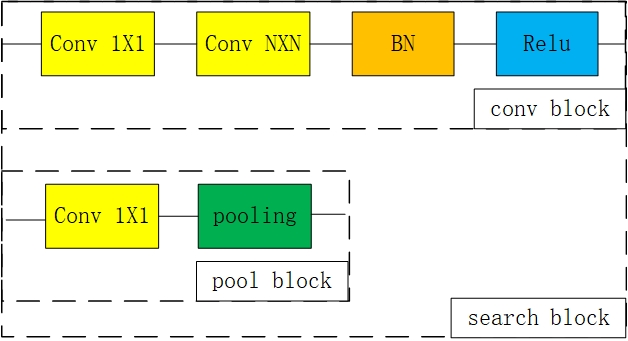}
	\end{center}
	\caption{The search block for NAS, which contains 3X3 depthwise separable convolution, 5X5 depthwise separable convolution, average pooling, and max-pooling}
	\label{search}
\end{figure}

For each mini-batch of data, LSTM fixes the policy to sample a network architecture. Momentum is served to optimize the parameters of the child network, by decaying the cross-entropy loss, which is the most commonly used in classification problems. To reduce the size of search space and the cost of training child networks, we draw on the known robust model, ENAS\cite{enas}, but we change the search space. Each node in the child network contains four operations, choosing from $3\times3$ or $5\times5$ separable convolution, average pooling, and max pooling. To save computational space and reduce inference latency, we replace all regular convolutions with depthwise separable convolutions in search space. As shown in Fig.\ref{search} these operations are encapsulated in blocks. The convolution blocks are connected in series by a $1\times1$ convolution, a $N\times N$ convolution, batch normalization, and RELU, where N and description are searched by NAS. At the same time, the pooling block is relatively simple with a $1\times1$ convolution added to max or average pooling to limit the channels of operations. Since the parameters are shared, the input and output shapes of each search block are fixed. This process goes through $Q$ iterations until all the input pictures are traversed, that represents the completion of an epoch for the child network.

\subsection{Training controller network}
After training the epoch of child networks, we fix child weights and combine different operations and connections to find the optimal architecture. The controller network is consist of an LSTM with 100 hidden units, whose gradient is updated by a reward signal. Reinforcement learning plays the role of an optimizer. We evaluate child network on each minibatch of the validation set to measure the accuracy, which is substituted into the reward function of reinforcement learning. As shown in the Eq\ref{reward}, we use coefficients to control the trade-off between network accuracy and complexity. Every time the child network infers on the verification set, it obtains not only the efficiency but also the inference latency \textit{L(m)} of the net. As expected, the inference latency raises with the increase of network complexity. So we added target latency to the reward function\cite{mnas}. At the same time, adding arcsin to the multi-objective reward function can give more positive feedback on the final small accuracy improvement.
\begin{equation} \label{reward}
\begin{aligned}
& \underset{m}{\text{maximize  }} & & arcsin({Accuarcy(m)} \times \left[ \frac{Latency(m)}{Target} \right] ^ q)
\end{aligned}
\end{equation} 
\noindent where $q$ is the weight factor.

This incentive mechanism reduces the size of the network to a great extent, in the final analysis, because the average verification accuracy and model complexity are balanced in the design of the reward function. To prevent over-fitting of the model, the controller generates different architectures each time. Finally, we summarize the whole NAS training procedure in Algorithm~\ref{enas}.

\begin{algorithm}[htb]
	\caption{Training Neural Architecture Search with Cross-Entry Loss}
	\label{enas}
	\begin{algorithmic}[1]
		\State  \textbf{Input:}{The training data$\{x_i,y_i\}_{i=1}^{n}$, the optimize function, the node numbers $L$, the controller train steps $S$, the quantity of sampled arc $Q$, the target latency $T$}
		\For {each epoch } {\do}
		\For {minibatch $B$ = $\{x_i,y_i\}_{i=1}^{B}$ in train$_{-}$dataset} {\do} 
		\State {controller(LSTM) sample architecture = "[arc]"}
		\State {child network update weights $W$ with "[arc]"}
		\EndFor
		\For {step in $S$}
		\State {Accuarcy = $ \frac{\sum_{i}^{w}eval_{-}acc}{Q} $ } \State {Latency = $ \frac{\sum_{i}^{w}eval_{-}lat}{Q} $ }
		\State {reward = $ arcsin({Accuarcy} \times \left[ \frac{Latency}{Target} \right] ^ q) $}
		\State {loss = F(reward)$\rightarrow$ controller update strategy $W_{controller}$}
		\EndFor
		\EndFor	
		\State  \textbf{Output:}{The network architecture}	
	\end{algorithmic}
\end{algorithm}

\subsection{Training fixed network}
At the end of the alternative training of child and controller network, we rank the architectures of the sample according to the accuracy of the verification set. The top three structures were taken out and trained from scratch. Data preprocessing includes clipping, flipping, cutout and so on, while momentum serves as a network optimizer. Each network is trained until converge, and its accuracy is verified on the test set. The algorithm flow is exhibited in Algorithm~\ref{fixed}

\begin{algorithm}[htb]
	\caption{Training fixed Neural Architecture from scratch}
	\label{fixed}
	\begin{algorithmic}[1]
		\State  \textbf{Input:}{The training data$\{x_i,y_i\}_{i=1}^{n}$, the fixed architecture $[selected$ $arc]$}
		\State controller(LSTM) build network with $[selected$ $arc]$
		\For {each epoch } {\do}
		\For {minibatch $B$ = $\{x_i,y_i\}_{i=1}^{B}$ in train$_{-}$dataset} {\do} 
		\State {child network update weights $W$ with "[selected arc]"}
		\EndFor
		\State {calculate accuary on the whole test dataset}
		\EndFor	
		\State  \textbf{Output:}{accuarcy of test dataset}	
	\end{algorithmic}
\end{algorithm}

\section{Experiments}
\label{Experiments}
We search large-scale data sets to find network structures and rank them according to the accuracy of network validation. Based on the hypothesis that the accuracy ranking of subnetworks is consistent with that of fixed networks, the top three networks are taken out and trained from scratch. Finally, we test the accuracy of the system on a small face data set.

\subsection{Face Data Set}
\subsubsection{Training Datasets}
\label{TrainingDatasets}
We employ the most generalized and challenging data set in face recognition (MS-Celeb-1M)\cite{msceleb} as a training set for the neural architecture search. The official MS-Celeb-1M data set consists of 10M pictures with 100k face identities, each of which has about 100 images. Because the original data set contains a lot of noise, we use a high-quality subset of MS-Celeb-1M improved by Arcface authors\cite{arcface}, which is optimized according to the distance from the image to the center of the class. The data set consists of 350K images, including 8.7k categories, with dozens to hundreds of faces in each category.

\subsubsection{Testing Dataset}
The testing dataset is LFW (Labeled Faces in the Wild)\cite{lfw}. LFW data set is established to study face recognition in unrestricted environments. This collection contains more than 13,000 face images (all from the Internet, not the laboratory environment). Every face is standardized by a person's name, among which about 1680 people have more than two faces. According to the standard LFW evaluation scheme, we measure the verification accuracy of 6000 faces. This dataset is widely used to evaluate the performance of face verification algorithm. Therefore, we use LFW to validate the performance of the network architecture retrieved by NAS.

\subsection{Training Details}

\subsubsection{Data preprocessing}
According to the preprocessing technology of face, we use MTCNN\cite{mtcnn} to detect five facial markers (eye center, nose tip, and mouth corner) for similarity transformation to align facial images. Each pixel in these images is normalized by subtracting the average of each channel and dividing by its standard deviation. At the same time, all the pictures are randomly flipped, padded to 120X120 and randomly cut to 112x112. Finally, the order of all data is shuffled. All of the above preprocessing is executed on both MS-Celeb-1M and LFW datasets. Also, we regularize the training set of MS-Celeb-1M with 16-size cutout to improve the robustness of the network. We divide MS-Celeb-1M data set into training, verification, and testing set according to 8:1:1.
\subsubsection{Network Settings}
\begin{figure}[t]
	\centering
	\label{some operations}	
	\subfigure[]{
		\includegraphics[width=0.45\linewidth]{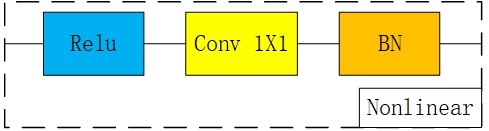}
		\label{unlinear}
	}
	\subfigure[]{
		\includegraphics[width=0.45\linewidth]{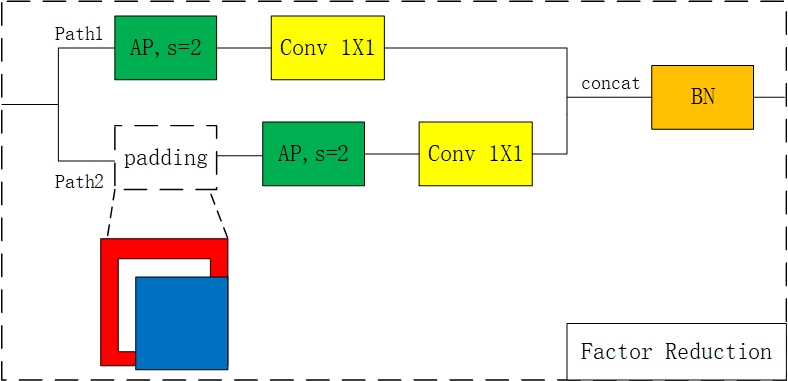}
		\label{factor}
	}
	\vspace{-1mm}
	\caption{The Fixed operation module in the NAS search processes (a). Nonlinear module: composed of  RELU function, $1\times1$ convolution and Batch Normalization (b). Factor Reduction module: halve the length and width of output images and double the number of channels}
\end{figure}

At the beginning of the network, we use a $3\times3$ convolution to extract features, followed by a Batch Normalization for regularization. As shown in Fig.\ref{unlinear}, the searched skip operations are processed by continuous RELU function, $1\times1$ convolution and Batch Normalization. This not only increases the nonlinearity of the neural network but also controls the number of channels in the transmission process. Between the first three search blocks in the network, we add a feature reduction module to each of them, in which the height and width of the image are halved, and the number of channels is doubled. We divide each module into two paths according to the number of channels. The first path passes through an average pooling with a stride of 2 and then follows a $1\times1$ convolution to fix the number of output channels. The second path is pre-processed especially: after a circle of padding around the image, the same size area of the original image in the lower right corner is exercised as a new image. Next up is the same path operation: an average pooling and a $1\times1$ convolution. We contact the two paths together as a new feature reduction operator. The detailed schematic diagram is exhibited in Fig.\ref{factor}. Considering the characteristics of parameter sharing, all skip operations need not only non-linear processing but also feature reduction at similar parallel positions. After the final search blocks, a drop block\cite{dropblock} and full connection layer are tailgated.

\subsubsection{Hyperparameter Selections}
We train NAS on an 8*V100 machine with a batch size of 128 per GPU. In the processing of training child networks, we adopt the most classical \textit{cross-entropy loss function} in face recognition. We set the number of search blocks on the child network to 5. To reduce the difference between independent training, we decrease the learning rate by cosine anneal.What's more, we add a function of training epochs\cite{sharp} to the original cosine annealing, as shown in Eq\ref{cosine}, which can converge to the optimal solution faster. To make the neural network better adapt to the tremendous learning rate, it warmups from 0 to 0.1 in the first 20 epochs and then reduces to 0.0001 according to the previous cosine annealing method. We use a weight decay of $10^{-4}$ and the optimizer of child network is momentum, with its $\beta$ value being fixed to 0.9.

\begin{equation}
\label{cosine}
lr_t = lr^i_{min} + (lr^i_{max} - lr^i_{min})\frac{2^{\frac{1}{2}
		\left(1 + \cos\left(\pi \frac{T_{cur}}{T_i}\right)\right) + 1}-2}{2}
\end{equation}
Where $lr^i_{min}$ and $lr^i_{max}$ are the minimum and maximum learning rates, respectively; $T_i$ is the total number of epochs; $T_{cur}$ is the current epoch; and $i$ is the index into a list of these parameters for a sequence of warm restarts in which $lr^i_{max}$ typically decays.

In contrast, the learning rate of the controller network is the strategy that piece-wise decreases from 0.1 to 0.0001 per 20 epochs. At the same time, the controller network uses the stochastic gradient descent (SGD) optimization algorithm. The child and controller networks alternately train 100 epochs, each of which traverses all training and validation data.

After the alternative training of child and controller networks, we rank the searched network architectures according to the verification accuracy. The top three structures are extracted and re-sampled with the controller network respectively. Besides increasing the total epochs from 100 to 150, all the hyperparameters of the fixed network are consistent with that of child network, including updating mechanism and optimizer of the learning rate, etc. At this time, all the data traversed by the former child and controller networks are regarded as the training data set. At the end of each epoch, we examine the accuracy on the last test set. All three fixed networks are randomly initialized and trained from scratch to convergence.

\begin{figure}[htbp]
	\centering	
	\subfigure[NAS A architecture]{
		\includegraphics[scale=0.3]{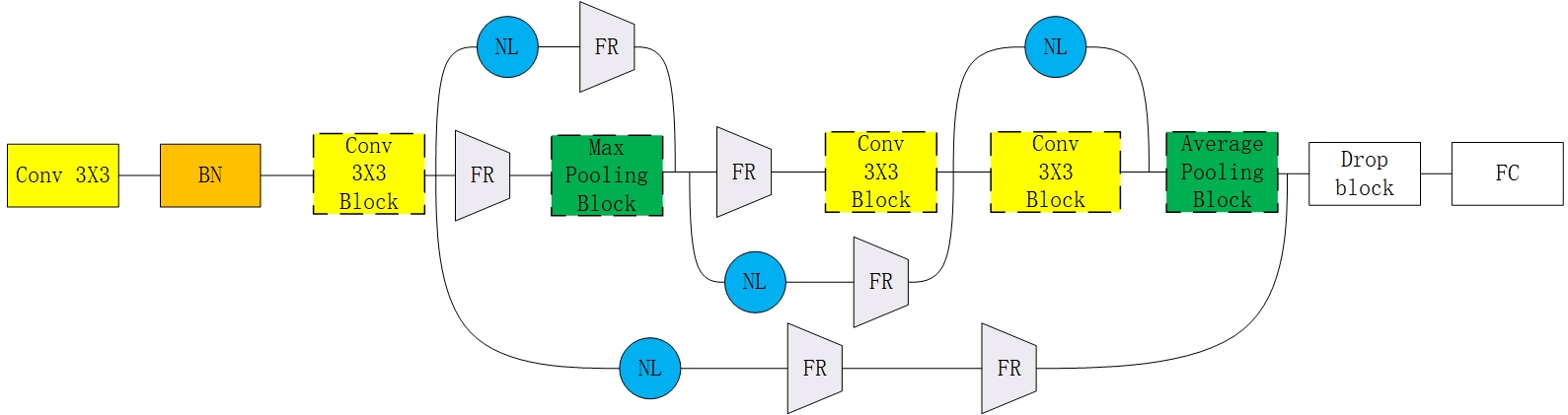}
		\label{arc1}
	}
	\subfigure[NAS B architecture]{
		\includegraphics[scale=0.3]{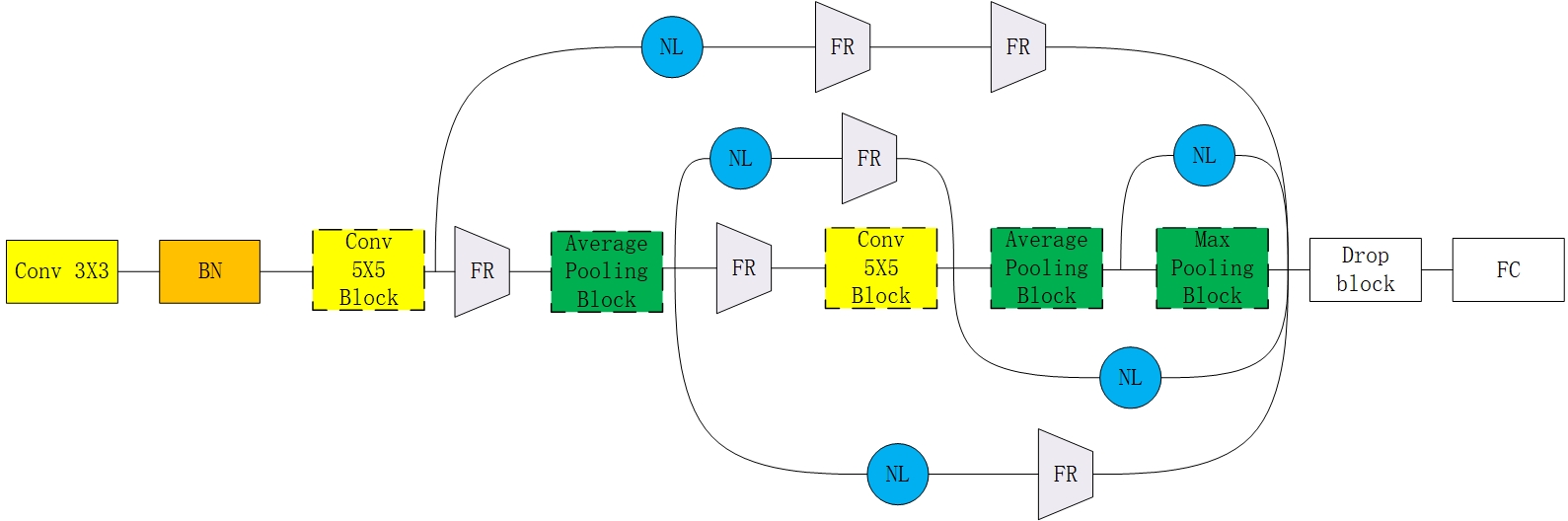}
		\label{arc2}
	}
	\subfigure[NAS C architecture]{
		\includegraphics[scale=0.3]{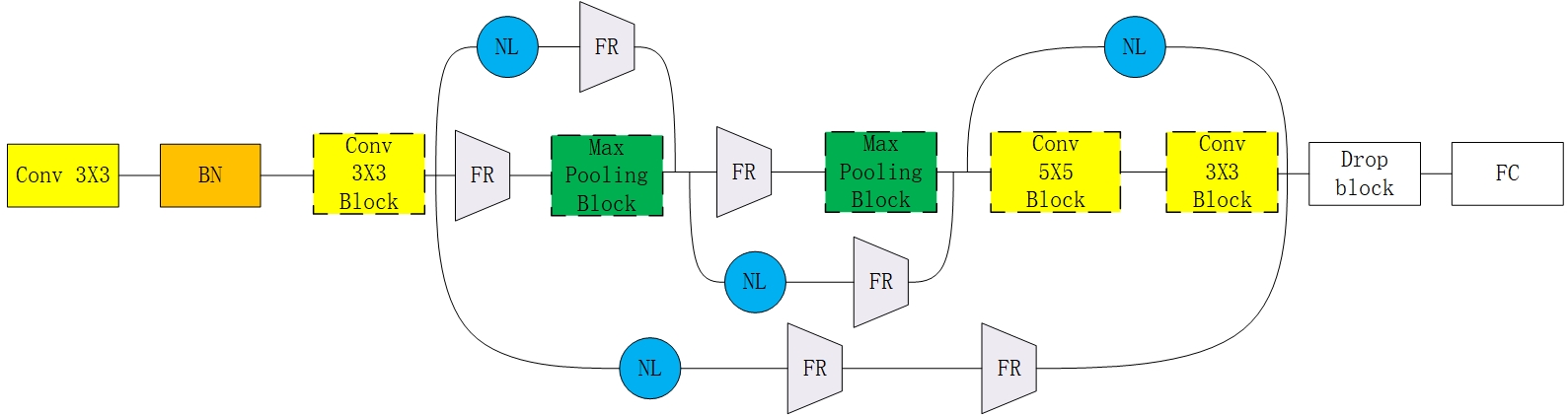}
		\label{arc3}
	}
	\vspace{-1mm}
	\caption{The final architectures of the top three fixed networks searched by NAS on MS-Celeb-1M data sets. The\textbf{ NL} and\textbf{ FR} represent the nonlinear and factor reduction module in the fixed operation module respectively, which have been described before. }
	\label{arc}	
\end{figure}

\subsection{Training Results}

MS-Celeb-1M data sets are divided into the batch size of 128, trained on 280W training sets and validated on 35W verification sets. 

After the training of child and controller networks is completed, the fixed networks are sorted by accuracy, and the top three are taken out and trained from scratch. In the network architecture, each skip operation is accompanied by the nonlinear module in Fig.\ref{unlinear}. At the same time, the first three blocks of the whole network and the corresponding location of the skip operations have the factor reduction module of Fig.\ref{factor}.

The schematic diagram of the network structure searched by NAS is shown in Fig.\ref{arc}. But the operators in search blocks and the connection of network architectures are different. After the fixed network training is completed, we test the accuracy of the network on the last 35W data sets, which is previously segmented in MS-Celeb-1M. We compared the test results of three networks with other previous studies in this data set.  As shown in Table.\ref{train}, all the NAS structures exceed previous results. Table.\ref{ms1m} shows the test accuracy of different methods on MS-Celeb-1M data sets. Among them, NASC achieved the state-of-art accuracy with \textbf{98.77\%}.

\begin{table}[htb]	  
	\begin{tabular*}{9.5cm}{lr}  
		\hline  
		Network  & Train Accuarcy(\%) \\  
		\hline 
		Resnet50 \& Cross Entropy\cite{compare} & 92.43 \\
		Resnet50 \& Angular Softmax & 93.33 \\
		Resnet50 \& Angular Margin Softmax & 93.68\\
		Resnet50 \& ArcFace & 92.34\\
		Resnet50 \& Marginal Loss & 91.57\\
		MobileNet v1 \& Cross ENtropy & 93.91 \\
		MobileNet v1 \& Angular Softmax & 93.45\\
		MobileNet v1 \& Angular Margin Softmax & 94.10\\
		MobileNet v1 \& ArcFace & 94.61\\
		MobileNet v1 \& Marginal Loss & 93.81\\
		\textbf{NAS A} \& Cross Entropy & \textbf{99.51} \\
		\textbf{NAS B} \& Cross Entropy & \textbf{99.63} \\
		\textbf{NAS C} \& Cross Entropy & \textbf{99.85} \\
		\hline  
	\end{tabular*} 	  
	\caption{The train accuracy of different networks with diverse loss functions on MS-Celeb-1M data sets.}
	\label{train}
\end{table}  

\begin{table}[htb]	  
	\begin{tabular*}{5.5cm}{lcr}  
		\hline  
		\textbf{Method} & \textbf{Test Accuarcy}(\%) \\  
		\hline  
		NAS A & 98.32 \\
		NAS B & 98.69 \\
		NAS C & 98.77 \\
		\hline  
	\end{tabular*} 	  
	\caption{The final test accuracy of the top three fixed networks searched by NAS on MS-Celeb-1M data sets. The test accuracy of NAS A reaches \textbf{98.32\%},  NAS B achieves \textbf{98.69\%}, and NAS C attains \textbf{98.77\%}}
	\label{ms1m}
\end{table}  

To further verify the generality of this method, we test it on LFW data set which is the most commonly used in face recognition. We fetch the three architectures and their parameters which had been trained by the previous fixed network and test them directly on LFW's 13,000 face images. As expected, NASB and NASC have achieved remarkable performance except for the slightly worse test results of NASA networks. Similar to the regular pattern of fixed network testing, NASC surpasses the existing face recognition technology with \textbf{99.89\%} accuracy and achieves the best result.

\begin{table}[htbp]	  
	\begin{tabular*}{8cm}{lcr}  
		\hline  
		Method & Image(M)  & Accuarcy(\%) \\  
		\hline  
		Deep Face\cite{deepface}  & 4.4 & 97.35 \\  
		Aug\cite{aug}  & 0.5 & 98.06 \\
		Fusion\cite{fusion}  & 500 & 98.37 \\
		MultiBatch\cite{Multibatch}  & 2.6 & 98.8 \\
		DeepFR\cite{parkhi}  & 2.6 & 98.95 \\
		Baidu\cite{targeting}  & 1.3 & 99.13 \\
		Center Loss\cite{centerloss}  & 0.7 & 99.28 \\
		LightCNN\cite{lightcnn}  & 4 & 99.33 \\
		SphereFace\cite{sphereface}  & 0.5 & 99.42 \\
		DeepID\cite{deepid2}  & 0.2 & 99.47 \\
		SphereFace+\cite{learning}  & 0.5 & 99.47 \\
		DeepID+\cite{DeepID2+}  & 0.3 & 99.47 \\
		Marginal Loss\cite{marginal}  & 3.8 & 99.48 \\
		MobileNet ArcFace Loss\cite{MobileNet}  & 5.8 & 99.50 \\
		Range Loss\cite{range}  & 5 & 99.52 \\
		DeepID3\cite{DeepID3}  & 0.3 & 99.52 \\
		MobileFaceNet\cite{MobileFaceNets}  & 3.8 & 99.55 \\
		FaceNet\cite{facenet}  & 200 & 99.63 \\
		CosFace\cite{CosineFace}  & 5 & 99.73 \\
		Resnet100 ArcFace Loss\cite{arcface}  & 5.8 & 99.83 \\
		NAS A   & 5.8 & 99.80 \\
		\textbf{NAS B}   & 5.8 & \textbf{99.84} \\
		\textbf{NAS C}   & 5.8 & \textbf{99.89} \\
		\hline  
	\end{tabular*} 
	\label{lfw} 
	\caption{The final test accuracy of the top three fixed networks searched by NAS on LFW data sets. The test accuracy of NAS A reaches 99.80\%,  NAS B achieves \textbf{99.84\%} and NAS C attains \textbf{99.89\%}}
\end{table}  

Although we consider the inference latency in Eq\ref{reward}, we set a relatively loose constraint range to improve the network test accuracy as much as possible. Interestingly, when the network is small, the speed of precision reduction is much slower than that of parameter reduction. But it is not that the more substantial of the system, the higher of the precision. Among the three network structures previously searched, NASB has the largest size, reaching 19.1M, but the highest efficiency is NASC, which has only 16M. Despite this, it is still better than traditional Resnet.This experimental result is helpful to select the appropriate architecture, which can obtain relatively small networks under the premise of excellent performance.

\section{Conclusions}
\label{Conclusions}
In the solution of the convolutional neural network, the efficiency of the model is mainly determined by the backbone network. In this paper, we change the traditional method of using the classical network as a backbone and modifying loss in face recognition with neural architecture search. To the best of our knowledge, this is the first time to search for a customized neural network in this field. The neural network we searched surpasses all previous results on MS-Celeb-1M dataset with 98.77\% accuracy. At the same time, we apply the search results directly to LFW data sets, and the test accuracy also reaches the state-of-art conclusion. The outstanding performance on these two famous face datasets proves the universality and excellence of our method.On the premise of guaranteeing the accuracy, we reduce the size of the searched network structure as much as possible.

Improving the efficiency of the NAS search and designing more general search space is still an unsolved problem, which we regard as our future work. At the same time, the combination of NAS and various loss functions will be a direction worthy of further exploration.

\section{Acknowledgement}

We would like to thank Huanyu Wang, Wenqi Fu, and Qirui Wang for their help and valuable advice in the development process.
\renewcommand\refname{7. Reference}

\bibliographystyle{unsrt}

\end{document}